\documentclass{article}
\usepackage{spconf,amsmath,graphicx}

\usepackage{enumitem}
\setlist{nosep, leftmargin=14pt}

\usepackage{mwe} 

\title{Low-Field Magnetic Resonance Image Quality Enhancement using a Conditional Flow Matching Model}
\name{Huu Tien Nguyen$^{(1)}$ and Ahmed Karam Eldaly$^{(1)(2)}$}
\address{$^{(1)}$Department of Computer Science, University of Exeter, Exeter, United Kingdom\\ $^{(2)}$UCL Hawkes Institute, University College London, London, United Kingdom}
\begin{document}
%
\maketitle
\begin{abstract}
This paper introduces a novel framework for image quality transfer based on conditional flow matching (CFM). Unlike conventional generative models that rely on iterative sampling or adversarial objectives, CFM learns a continuous flow between a noise distribution and target data distributions through the direct regression of an optimal velocity field. We evaluate this approach in the context of low-field magnetic resonance imaging (LF-MRI), a rapidly emerging modality that offers affordable and portable scanning but suffers from inherently low signal-to-noise ratio and reduced diagnostic quality. Our framework is designed to reconstruct high-field–like MR images from their corresponding low-field inputs, thereby bridging the quality gap without requiring expensive infrastructure. Experiments demonstrate that CFM not only achieves state-of-the-art performance, but also generalises robustly to both in-distribution and out-of-distribution data. Importantly, it does so while utilising significantly fewer parameters than competing deep learning methods. These results underline the potential of CFM as a powerful and scalable tool for MRI reconstruction, particularly in resource-limited clinical environments.
\end{abstract}

\vspace{-0.1cm}
\section{Introduction}
\label{sec:intro}
\vspace{-0.1cm}
Magnetic resonance imaging (MRI) is a cornerstone of radiology, but image quality is highly dependent on field strength: high-field systems ($>1$T) provide superior resolution and contrast, whereas low-field scanners ($<1$T) suffer from reduced signal-to-noise ratio and artifacts that limit diagnostic accuracy. There is a pressing need for cost-effective methods that deliver high-field-like quality \cite{murali2024bringing}. To address this challenge, various image enhancement methods have been proposed. Super-resolution approaches attempt to improve spatial resolution by upsampling low-resolution scans, while modality transfer methods consider the problem as a cross-domain translation task, learning mappings across imaging modalities (e.g., CT-to-MR synthesis) to recover richer information \cite{gu2023cross}. Another line of work, image quality transfer (IQT), leverages high-quality data from advanced imaging systems to improve routinely acquired scans. Early IQT approaches employed patch-based random forests \cite{alexander2014image}, which captured nonlinear relationships between low- and high-quality image patches. More recently, Lin et al. \cite{lin2023low} introduced stochastic simulation combined with a directional U-Net for low-field MRI enhancement, showing further improvements.

Generative modeling has driven additional progress. Diffusion models have emerged as powerful frameworks for image enhancement \cite{kim20233d}. In parallel, unsupervised and hybrid dictionary learning approaches have demonstrated improved robustness to out-of-distribution data, an essential property given the variability of clinical imaging \cite{eldaly2024alternative}. These studies highlight the potential of generative and representation learning methods, yet challenges remain. Existing techniques often rely on large model sizes, iterative sampling procedures, or struggle to generalise reliably across diverse acquisition settings. Thus, in this work, we introduce conditional flow matching (CFM) — a recent generative paradigm that learns continuous flows between noise and data distributions through velocity field regression \cite{lipman2022flow}. Unlike diffusion models, CFM achieves this without iterative denoising, offering a more efficient and compact alternative. To our knowledge, this is the first application of CFM to IQT for low-field MRI enhancement. The main contributions of this work are as follows. (1) We propose a novel CFM-based framework for reconstructing high-field-like images from low-field MRI. (2) We provide a comprehensive benchmarking against state-of-the-art IQT approaches, demonstrating that CFM achieves superior quality while using significantly fewer parameters, with robust performance in both in-distribution and out-of-distribution datasets.

\vspace{-0.1cm}
\section{Methods}
\label{sec:format}
\vspace{-0.1cm}
\subsection{Conditional Flow Matching process}
\begin{figure}
\centering	
\includegraphics[width=\linewidth]{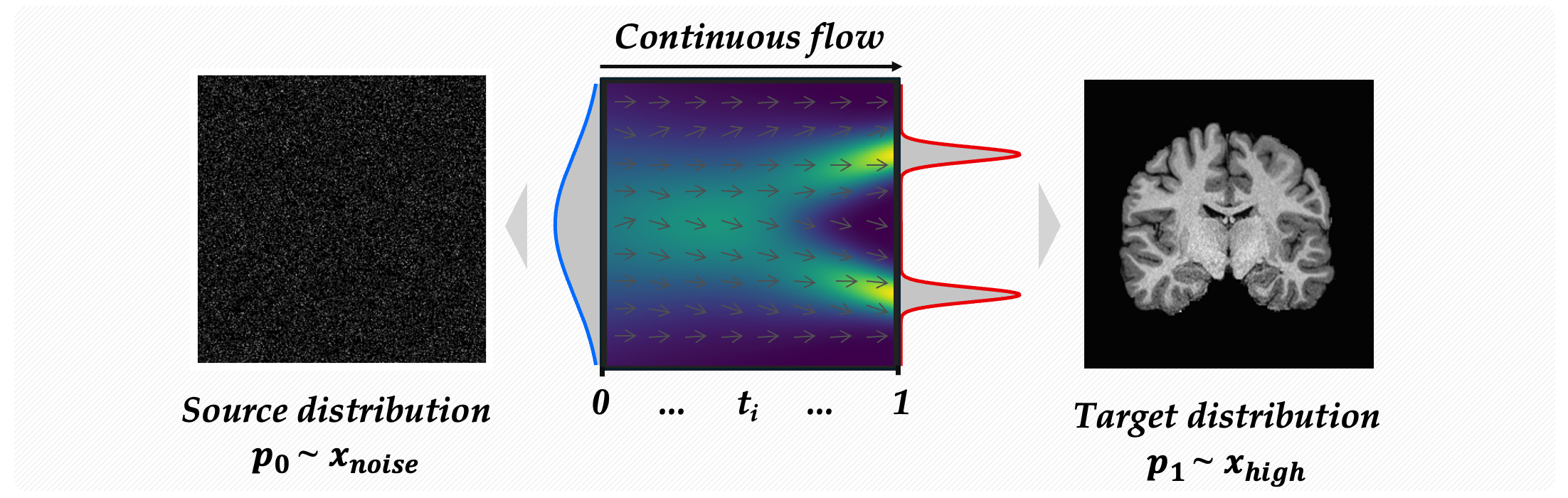}
\caption{CFM learns velocity fields between noise and data distributions, then generates samples by following these fields.}
\label{image_continuous_flow_paper_isbi}
\end{figure}

\vspace{-0.1cm}
CFM assumes a continuous transformation from a source distribution \( p_0 \sim x_{\text{noise}} \) to a target distribution \( p_1 \sim x_{\text{high}} \) over a time interval \( t \in [0, 1] \), as shown in Fig. \ref{image_continuous_flow_paper_isbi}. This transformation can be modeled as an ordinary differential equation \eqref{ode_fc} where \( x_t \) is the intermediate image at time \( t \). as follows
\begin{equation}
\frac{d x_t}{d t} = v_\theta(x_t, t \mid x_{\text{low}}). \quad \label{ode_fc}
\end{equation}

After learning the velocity field $f_\theta(x_t, t \mathrel{|} x_{low})$
, we can sample by solving the differential equation \eqref{ode_fc}. A common method to solve this equation is the Euler method as in Eq. \eqref{euler_method}, and the final result is an output sample.
\begin{equation}
x_{noise}(t + \Delta t) \approx x_{noise}(t) + f_\theta(x_{noise}, t | x_{low})\Delta t.
\label{euler_method}
\end{equation}

\vspace{-0.1cm}
\subsection{Network architecture}
\begin{figure}
\centering	
\includegraphics[width=\linewidth]{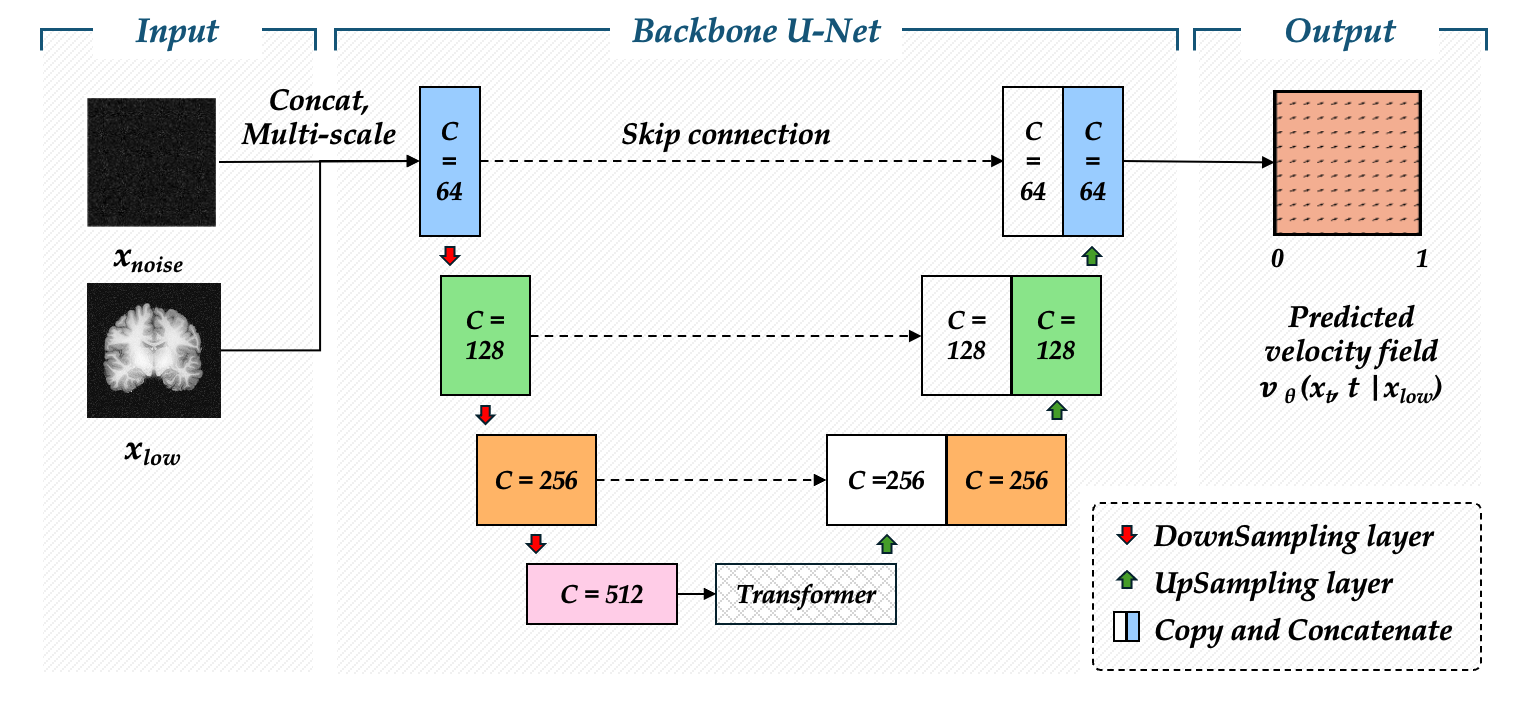}
\caption{Integration of CFM with U-Net backbone: multi-scale inputs, residual blocks with channel attention, transformer-enhanced bottleneck, and advanced sampling methods.}
\label{fig:image_unet_isbi_1c}
\end{figure}

\vspace{-0.1cm}
Figure \ref{fig:image_unet_isbi_1c} shows network architecture, which is built upon the foundation of U-Net \cite{ronneberger2015u}. A distinctive feature of U-Net is the symmetric skip connections between encoder and decoder layers, enabling direct concatenation of high-resolution feature maps. This mechanism combines abstract semantic information from deep layers with precise spatial information from shallow layers, crucial for tasks requiring high positional accuracy and image reconstruction. Based on this foundation, we propose a customised U-Net architecture for IQT using CFM, enhanced with several components as follows.

\noindent\textit{Multi-scale input layer:} The network input is a tensor concatenated from the intermediate image $(x_t)$ and the conditional image $(x_{low})$. To effectively extract initial features, the first convolutional layer of traditional U-Net has been replaced with a multi-scale input layer, where we process it in parallel through multiple convolutional branches with different kernel sizes ($1\times 1$, $3\times 3$, $7\times 7$, and $15\times 15$). The output from each branch is then concatenated along the channel dimension to create a rich input feature map for the first encoder layer. This approach allows the model to simultaneously capture both fine details through small kernels and larger structures through large kernels right from the first layer rather than having to wait until deeper layers in the network.

\begin{figure}
\centering	
\includegraphics[width=\linewidth]{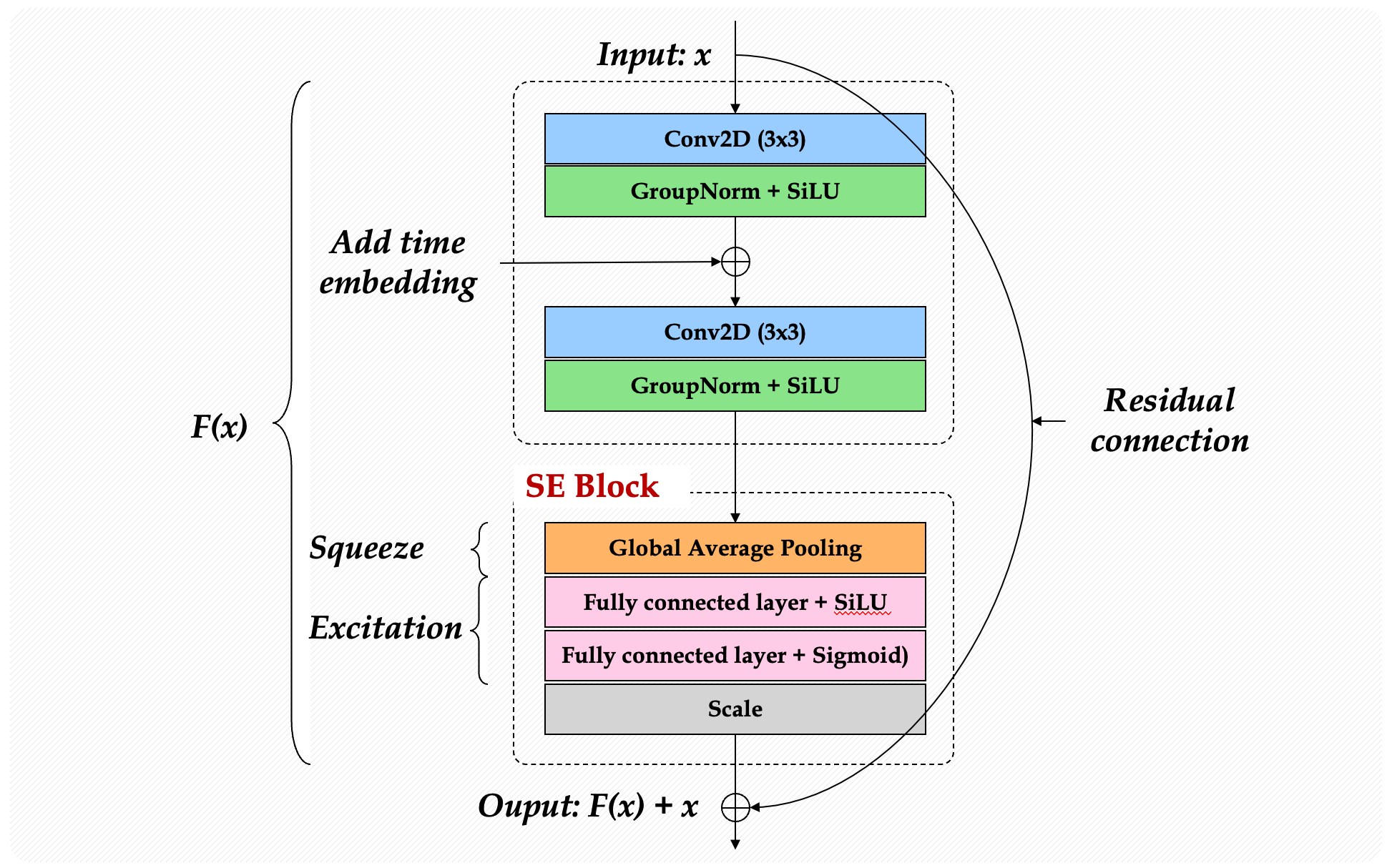}
\caption{ResnetBlock serves as the core feature processing component in both encoder and decoder.}
\label{image_resnet_isbi}
\end{figure}

\noindent\textit{Residual block with Squeeze-and-Excitation (SE) module}: As shown in Fig.\ref{image_resnet_isbi}, each Resnet block consists of two $3\times 3$ convolutional layers (Conv2d), with each layer followed by a group normalisation layer (GroupNorm) and a SiLU activation function. To condition the processing according to the time variable $t$ of the CFM method, a time embedding vector is incorporated into the block. This vector is first processed by a small multi-layer perceptron network and then added to the feature map after the first convolutional layer. Additionally, to enhance representational capacity, each block is also integrated with a SE module. This module implements a channel attention mechanism, allowing the model to automatically learn how to adjust the importance of each feature channel \cite{hu2018squeeze}. 

\noindent\textit{Pixel unshuffle for down-sampling and pixel shuffle for up-sampling}: To change spatial resolution, our model employs symmetric pixel unshuffle and pixel shuffle operations instead of conventional max-pooling or ConvTranspose2d layers. During downsampling, pixel unshuffle (space-to-depth) rearranges $2\times 2$ pixel blocks from spatial dimensions into channels, quadrupling channel count while halving spatial dimensions, followed by $1\times 1$ convolution to adjust channels for the next layer. Conversely, during up-sampling, we use pixel shuffle using a $1\times 1$ convolution quadruples channels, then Pixel Shuffle (Depth-to-Space) spreads these channels across spatial dimensions, doubling height and width while restoring original channel count.

\noindent\textit{Bottleneck transformer block} 
The bottleneck contains a transformer block allowing multi-head self-attention mechanisms to be applied in this deeper feature map. This allows the model to more effectively identify and enhance relationships between distant image regions before the reconstruction process begins in the decoder part.

\vspace{-0.1cm}
\section{Experiments and results}
\vspace{-0.1cm}
\subsection{Datasets}
\vspace{-0.1cm}
We use high-resolution T1-weighted (T1w) MR images from the Human Connectome Project (HCP) \cite{sotiropoulos2013advances}. The original brain images were acquired on a 3 Tesla Siemens Connectome scanner, with parameters of 0.7-mm isotropic voxel, repetition time (TR) of 2400 ms, echo time (TE) of 2.14 ms, and inversion time (TI) of 1000 ms for T1w images. To generate training pairs from T1w MRI images, we apply a stochastic low-field image simulator \cite{lin2023low} that transforms high-quality images into low-field counterparts with realistic noise and contrast variations. The InD datasets used for training and testing are synthesised with parameters constrained by a Mahalanobis distance $< 1$, ensuring that the SNR in white matter (WM) is higher than in gray matter (GM) to maintain tissue contrast compatible with T1-weighted images. On the other hand, the OOD test dataset is synthesised by sampling SNR parameters from a distribution reflecting ultra-low field T1-weighted MRI images, introducing significant deviations from the Gaussian distribution used for the InD data for training. For training, 404 image pairs were randomly split: 80\% (323 pairs) for training and 20\% (81 pairs) for validation. For testing, we use two evaluation sets: an InD test set with 300 pairs sharing training distribution characteristics, and an OOD test set with 200 pairs having different distribution properties to evaluate generalisation capability and robustness in real-world applications where new data differs from training data.

\vspace{-0.1cm}
\subsection{Implementation details}
\vspace{-0.1cm}
All training and testing procedures were implemented in Python 3.9.18 and executed on an NVIDIA H100 NVL GPU system. The IQT-CFM model was trained using the Adam optimizer with an initial learning rate of $10^{-4}$ and L2 weight decay of $10^{-4}$ to prevent overfitting. Training was conducted with a batch size of 4 over 200 epochs to ensure loss function convergence. Additionally, we employed a Cosine Annealing learning rate scheduler that gradually decreases the learning rate following a sinusoidal cycle from the initial value of $10^{-4}$ to a minimum of $10^{-6}$. The trained model weights obtained from training process were subsequently used during the inference phase to evaluate performance on both InD and OOD test datasets.

\vspace{-0.1cm}
\subsection{Comparison with other work and evaluation criteria}
\vspace{-0.1cm}
The proposed algorithm is compared to the baseline interpolation method which uses a stochastic low-field image simulator \cite{lin2023low} to generate low-field versions from high-quality images, and the state-of-the-art deep dictionary learning IQT method, (IQT-DDL) \cite{eldaly2024alternative}. The performance of the proposed approach and the existing methods is assessed using peak signal-to-noise ratio (PSNR), structural similarity index measure (SSIM), and learned perceptual image patch similarity (LPIPS).

\vspace{-0.1cm}
\subsection{Quantitative results}
\begin{sloppypar}
\vspace{-0.1cm}
Table \ref{tab:compare_metrics_ind_ood} presents the quantitative evaluation results on both in-distribution (InD) and out-of-distribution (OOD) datasets. 
On the InD dataset, the proposed IQT-CFM approach consistently achieved superior performance across all evaluation metrics. It delivered a noticeable improvement in reconstruction fidelity compared with both the state-of-the-art IQT-DDL model and the interpolation baseline, leading to enhanced structural preservation and perceptual quality. Furthermore, these gains were achieved with substantially fewer learnable parameters, highlighting the efficiency of the proposed framework. 
A similar trend was observed on the OOD dataset, where IQT-CFM continued to outperform the competing methods in terms of reconstruction quality and perceptual similarity, despite the expected drop in absolute performance across all approaches. The method maintained a modest yet consistent advantage, indicating stronger generalisation beyond the training distribution. Overall, these results demonstrate that IQT-CFM not only achieves state-of-the-art image reconstruction quality but also does so with significantly improved parameter efficiency.
\end{sloppypar}

\begin{table}
\caption{Quantitative results using InD and OOD datasets. Best results indicated in bold.}
\label{tab:compare_metrics_ind_ood}
\centering
\small
\setlength{\tabcolsep}{4pt}
\begin{tabular}{@{}lrrr@{}}
\hline
\textbf{Metric} & \textbf{Interpolation} & \textbf{IQT-DDL} & \textbf{IQT-CFM} \\
\hline\hline
\textbf{InD Dataset} \\
\hline
PSNR$\uparrow$     & $28.92\pm0.443$ & $36.07\pm0.903$ &  \textbf{37.07$\pm$1.024} \\
SSIM$\uparrow$     & $0.88\pm0.013$  & $0.95\pm0.009$  & \textbf{0.96$\pm$0.007}  \\
LPIPS$\downarrow$    & $0.12\pm0.011$  & $0.05\pm0.010$  & \textbf{0.03$\pm$0.008}  \\
Params$\downarrow$   & N/A             & 9,337,203       & \textbf{5,253,249}  \\
\hline\hline
\textbf{OOD Dataset} \\
\hline
PSNR$\uparrow$     & $24.69\pm0.602$ & $26.25\pm0.823$ & \textbf{26.33$\pm$0.821} \\
SSIM$\uparrow$     & $0.76\pm0.050$  & $0.76\pm0.061$  & \textbf{0.77$\pm$0.050}  \\
LPIPS$\downarrow$    & $0.16\pm0.061$  & $0.14\pm0.065$  & \textbf{0.13$\pm$0.062}  \\
Params$\downarrow$   & N/A             & 9,337,203       & \textbf{5,253,249}  \\
\hline
\end{tabular}
\end{table}

\subsection{Qualitative results}
\begin{figure}[ht]
\centering	
\includegraphics[width=\linewidth]{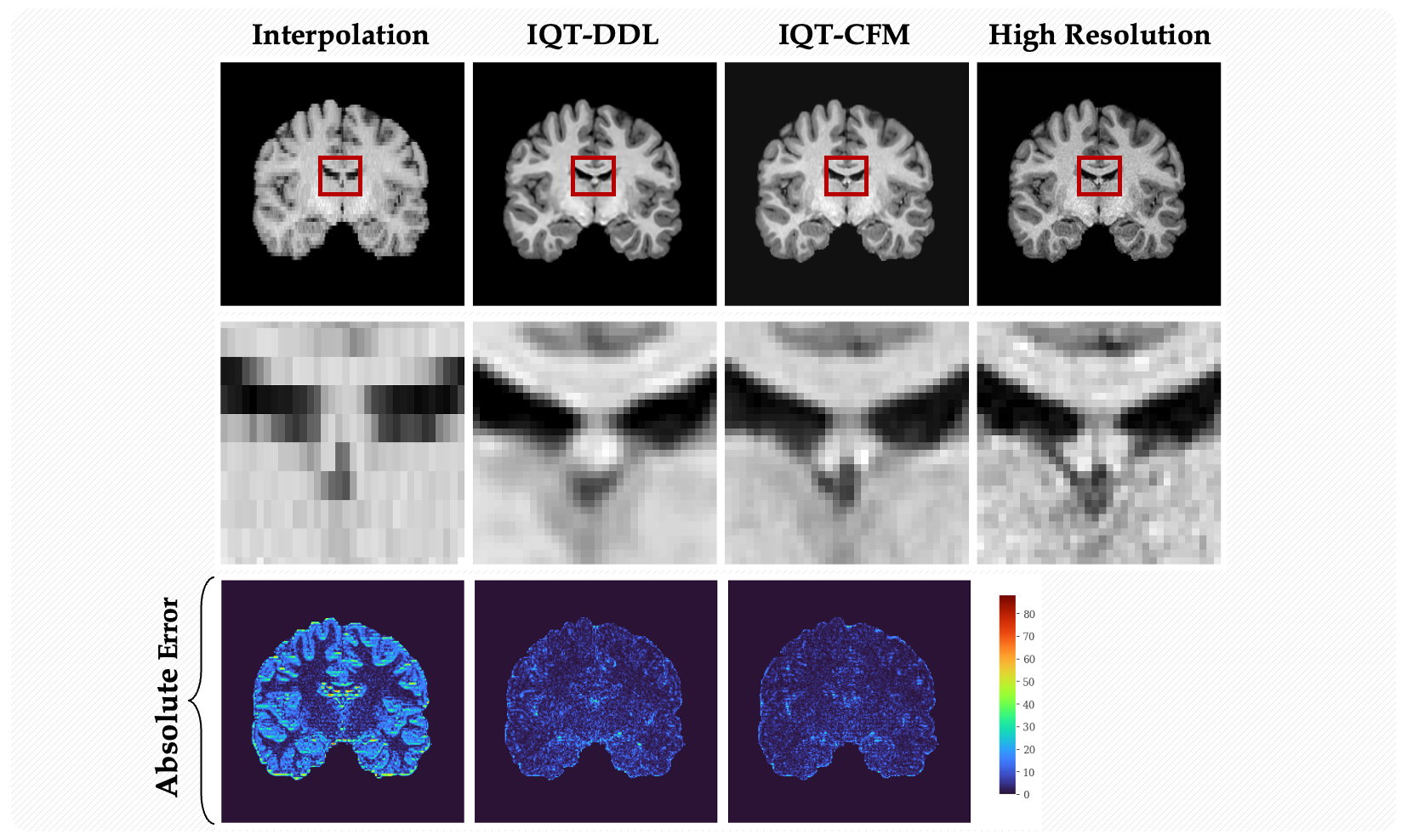}
\vspace{-0.1cm}
\caption{Example of enhanced brain image using InD data and the corresponding error maps.}
\label{image_ind_overall_isbi}
\vspace{-0.1cm}
\end{figure}

\begin{figure}[ht]
\centering	
\includegraphics[width=\linewidth]{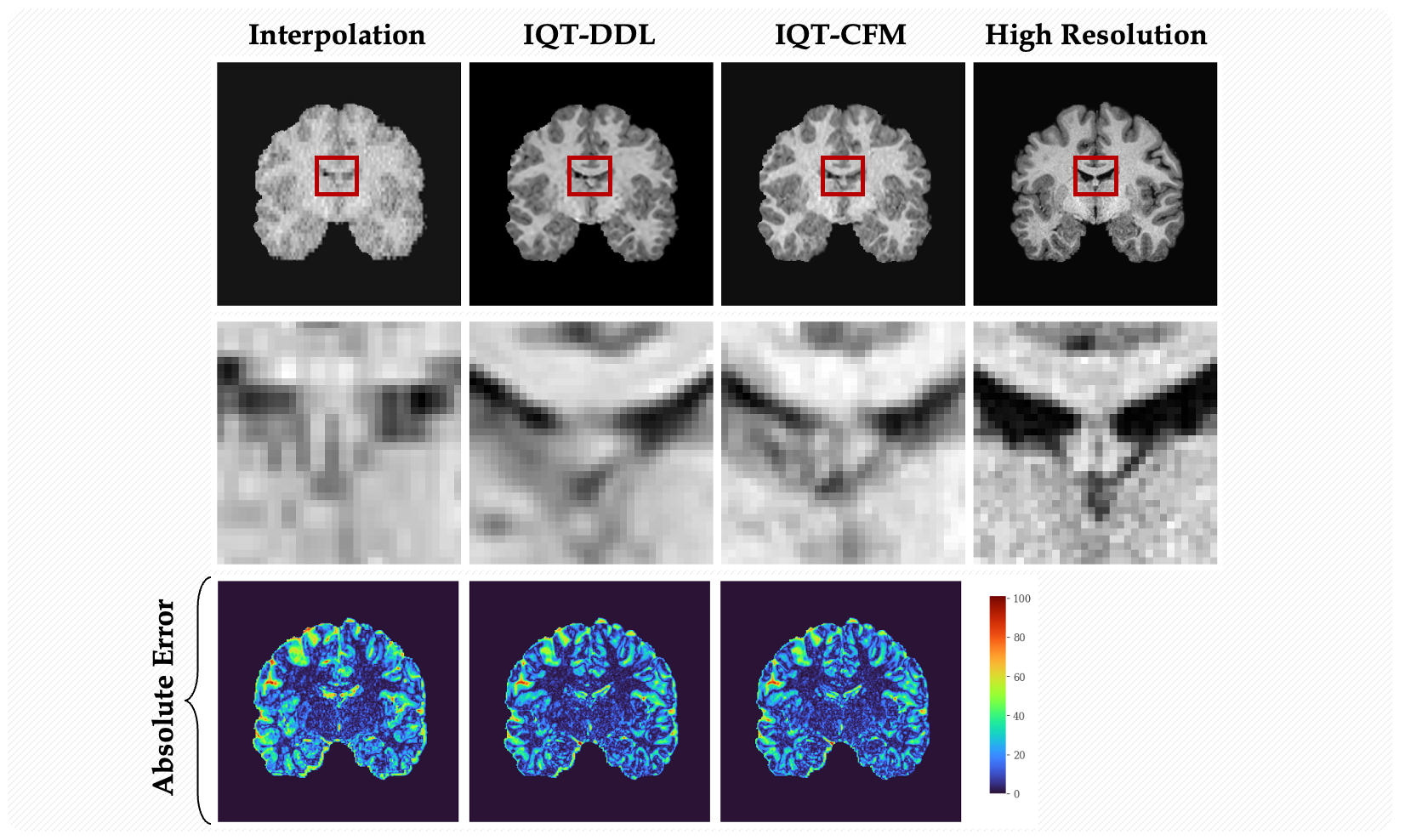}
\vspace{-0.1cm}
\caption{Example of enhanced brain image using OOD data and the corresponding error maps.}
\label{image_ood_overall_isbi}
\vspace{-0.1cm}
\end{figure}

\vspace{-0.1cm}
Figures~\ref{image_ind_overall_isbi} and~\ref{image_ood_overall_isbi} show reconstructed brain images from the HCP dataset using the InD and OOD test sets, along with the corresponding error maps that illustrate the absolute differences between the ground-truth high-resolution images and their reconstructions. The qualitative results are consistent with the quantitative findings in Table \ref{tab:compare_metrics_ind_ood}. As expected, interpolation produces noticeable blurring and a loss of structural details, whereas IQT-CFM achieves best reconstructions with substantially lower errors compared to both interpolation and IQT-DDL. Furthermore, error maps derived from the OOD dataset reveal higher residuals than those from the InD dataset. Such performance reflects the common generalisation challenge of domain shifted data, but simultaneously emphasises the importance of evaluating model generalisation.

\vspace{-0.1cm}
\section{Conclusions}
\begin{sloppypar}
\vspace{-0.1cm}
In this work, we introduced a novel conditional flow matching (CFM) framework for image quality transfer (IQT) for low-field MRI quality enhancement. By learning continuous flows between noise and data distributions, CFM provides an efficient alternative to conventional generative approaches. Comprehensive evaluations using both in-distribution and out-of-distribution datasets demonstrated that IQT-CFM achieves state-of-the-art reconstruction quality with significantly fewer parameters than existing IQT methods. This efficiency, coupled with robust generalisation, makes CFM particularly well-suited for clinical deployment in resource-limited settings, where access to high-field scanners is restricted. Future work will explore integrating uncertainty estimation for improved clinical reliability.
\end{sloppypar}

\vspace{-0.1cm}
\section{COMPLIANCE WITH ETHICAL STANDARDS}
\vspace{-0.1cm}
This research study was conducted retrospectively using a publicly available brain MRI scan dataset \cite{sotiropoulos2013advances}. Ethical approval was not required.

\bibliographystyle{IEEEbib}
\bibliography{strings,refs}

\end{document}